\documentclass{article}

     \usepackage[final]{neurips_2019}

\usepackage[utf8]{inputenc} 
\usepackage[T1]{fontenc}    
\usepackage{hyperref}       
\usepackage{url}            
\usepackage{booktabs}       
\usepackage{amsfonts}       
\usepackage{nicefrac}       
\usepackage{microtype}      
\usepackage{graphicx}
\usepackage{multirow}
\usepackage{xspace}
\usepackage{subcaption}

    \title{NLVR2 Visual Bias Analysis}
    
    \author{Alane Suhr and Yoav Artzi\\ Cornell University \\ {\tt \{suhr, yoav\}@cs.cornell.edu }}
    
    \date{}

\newcommand{\true}{\texttt{true}\xspace}
\newcommand{\false}{\texttt{false}\xspace}

\begin{document}

    \maketitle

\begin{abstract}
NLVR2~\citep{Suhr2019:nlvr2} was
designed to be robust for language bias through a data collection
process that resulted in each natural language sentence appearing with
both true and false labels. The process did not provide a similar
measure of control for visual bias. This technical report analyzes the potential for
visual bias in NLVR2. We show that some amount of visual bias likely
exists. Finally, we identify a subset of the test data that allows to
test for model performance in a way that is robust to such potential
biases. We show that the performance of existing models \citep{Li:19visualbert,Tan:19lxmert} is relatively robust to this potential bias. We
propose to add the evaluation on this subset of the data to the NLVR2
evaluation protocol, and update the official release to include it. 
A notebook including an implementation of the code used to replicate this analysis
is available at \url{http://nlvr.ai/NLVR2BiasAnalysis.html}.
\end{abstract}

\section{Introduction}\label{sec:intro}

NLVR~\citep{Suhr:17visual-reason} and NLVR2~\cite{Suhr2019:nlvr2} are two recently proposed benchmarks for visual reasoning with natural language. Both uses a simple binary classification task:
given an image and a natural language statement, predict if the
statement is \true or \false with regard to the image.

\begin{figure}[h]
    \centering
\begin{subfigure}{0.49\columnwidth}
\centering
    \fbox{\includegraphics[width=\textwidth]{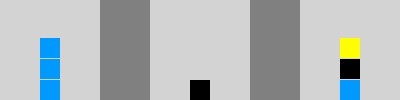}} \\[5pt]
    \fbox{\includegraphics[width=\textwidth]{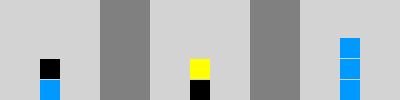}} \\[5pt]
    \fbox{\includegraphics[width=\textwidth]{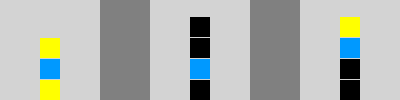}} \\[5pt]
    \fbox{\includegraphics[width=\textwidth]{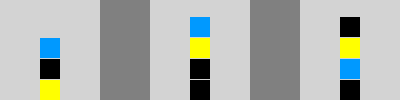}}
    \caption{Examples from NLVR~\citep{Suhr:17visual-reason}. All examples are for the statement: \textit{there is a yellow block as the top of a tower with exactly three blocks}. The labels are, from top to bottom: \true, \false, \true, \false.}
    \label{fig:nlvr}
    \end{subfigure}
    \hfill
 \begin{subfigure}{0.4\columnwidth}
    \centering
    \fbox{\includegraphics[width=\textwidth]{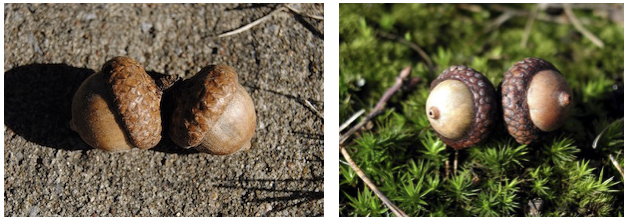}} \\[5pt]
    \fbox{\includegraphics[width=\textwidth]{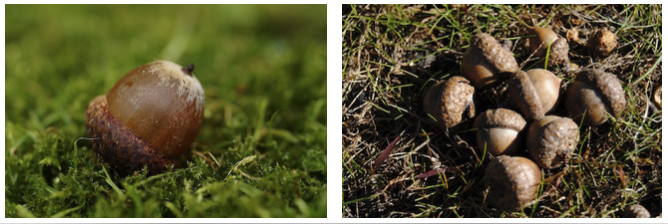}} \\[5pt]
    \fbox{\includegraphics[width=\textwidth]{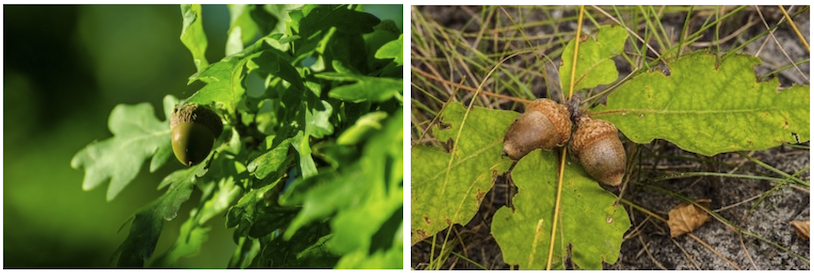}} \\[5pt]
    \fbox{\includegraphics[width=\textwidth]{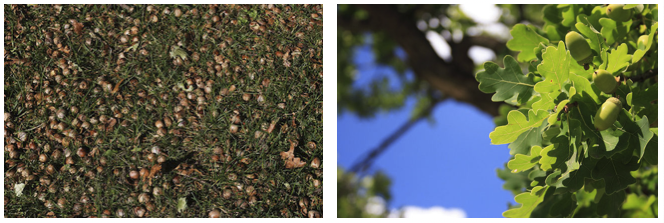}}
\caption{Examples from NLVR2~\citep{Suhr2019:nlvr2}. All examples are for the statement: \textit{One image shows exactly two brown acorns in back-to-back caps on green foliage.}. The labels are, from top to bottom: \true, \false, \true, \false.}
    \label{fig:nlvr2_examples}
    \end{subfigure}
    \caption{Examples from NLVR and NLVR2.}
\end{figure}

The images in NLVR are synthetically generated, and the language was
collected through a crowdsourcing process. The image generation and data
collection process was designed to generate data robust to
single-modality biases. Each statement is paired with multiple images,
some pairings have the label \true, while others are
\false. The images were generated so that for each image used in
an example with the label \true, there is an image with exactly
the same set of objects but arranged differently in an example with the
same sentence, but with the label \false. Figure~\ref{fig:nlvr} shows four
examples from NLVR, all using the same statement.

NLVR2~\citep{Suhr2019:nlvr2} uses  two paired real images
and a natural language statement for a similar classification setup, where the goal is to decide if the statement
is \texttt{true} or \texttt{false} with the regard to the pair of
images. The NLVR2 data collection process included creating a new set of
web images and a compare and contrast sentence-writing task, which
enabled collecting semantically diverse natural language.

\section{Overview of the NLVR2 Data Collection Process} 

The data collection
process is split into: (a) collecting sets of image pairs, (b) setntence
writing, and (c) validation. In the first step, sets of eight related
images are collected and randomly split into four pairs. Each set of
four pairs is given to a worker. They were asked to select two pairs,
and write a sentence that is \texttt{true} for the selected pairs and
\texttt{false} for the other pairs. Finally, workers are presented with
an image pair and a statement to validate the label assigned by the
sentence writer. This leads to pruning a small part of the data, so not
all statements appear exactly four times (i.e., with the four orignal
image pairs). Figure~\ref{fig:nlvr2_examples} shows an example statement from NLVR2 paired
with four image pairs, two with a label of \texttt{true} and two with
\texttt{false}.\footnote{License information for images in Figure~\ref{fig:nlvr2_examples}, from top left to bottom right: Charles Rondeau (CC0), Hagerty Ryan, USFWS (CC0), Andale (CC0), Charles Rondeau (CC0), George Hodan (CC0), Maksym Pyrizhok (PDP), Peter Griffin (CC0), Petr Kratochvil (CC0).}

\begin{figure}[t]
\begin{center}
\includegraphics[width=0.5\textwidth]{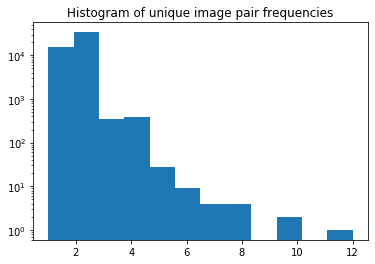}
\end{center}
\caption{Histogram demonstrating the frequency of unique image pairs appearing in the training set. The bins (x-axis) represent frequencies that a unique pair appears in the data, and the y-axis shows how many unique pairs appear with that frequency.}
\label{fig:uniq_freq}
\end{figure}

During the annotation process,
pairs of images are often annotated multiple times, and appear in the
data with different sentences. The majority of such cases is because
each set of pairs was annotated twice to obtain more natural language
data with the available of sets of image pairs. However, a small number
of pairs were annotated even more times because they appeared in
multiple sets (i.e., set of four pairs). This happened by chance, and
relatively rarely. Figure~\ref{fig:uniq_freq} shows the distribution of
times unique image pairs appear in the training set.


\section{Visual Bias in NLVR2}\label{sec:bias}

Each sentence in NLVR2 is written to be
\texttt{true} for two pairs of images and \texttt{false} for two pairs.
This way, each sentence appears with multiple image pairs with different
labels (unless some were pruned in validation). This provides robustness
against language biases (i.e., some sentences are more likely to be
\texttt{true} or \texttt{false} depending on the language only).
However, it is possible that some image
pairs are more likely to be picked so that the sentence written for them
has the label \texttt{true} and some are more likely to elicit the label
\texttt{false} based on the visual content alone. This can lead to
models that use this bias to solve the task while ignoring the natural
language statement. This may happen because it was left to the workers
to pick which pairs to label as \texttt{true} and \texttt{false}. This choice was left to the workers  to make the task easier. \citet{Suhr2019:nlvr2} note
that forcing the label assignment creates a much harder task of what is
already a challenging annotation task.
We analyze this potential for bias using pairs that appear in the data multiple times.

Consider pairs that occur twice in the data. 
If labels are assigned randomly by the
sentence writers, we would expect about half of these pairs to have
exactly the same label in both instances (\texttt{true}/\texttt{true} or
\texttt{false}/\texttt{false}). In total, there are 33,866 pairs in the training set that occur twice,
so we expect 16,933 pairs to have the same label in both instances.
However, the observed frequency of pairs with the same label is higher, at 21,267.
The pairs that occur twice likely appeared along with the same images
during sentence-writing. Workers were prompted to write a sentence true
about the selected pairs and false about the unselected pairs, and
therefore are likely to prefer to select the two pairs that are more
similar to each other within the set of four pairs. This could explain a
bias towards a certain label for that pair. Therefore, it is likely  that the bias exists only when considering each image pair in
the context of the other pairs it appeared with in sentence-writing.
When taken out of this context, as in the NLVR2 task, this bias may
become irrelelvant.

Pairs that occur more than twice in the data are more interesting.
Consider an event that is less likely: pairs that occur three times with
the same label. The probability for a pair to have the same label
(\texttt{true}/\texttt{true}/\texttt{true} or
\texttt{false}/\texttt{false}/\texttt{false}) is 0.25.
There are 345 pairs that occur three times, so we would expect around 43 to have the same label in all instances.
However, we observe 128 have the same label.
We can generalize this analysis to all observed pair frequencies in the
training data. Table~\ref{tab:freqs} shows the expected and observed counts of unique pairs appearing with the same label
across instances for varying pair frequencies.

\begin{table}[t]
\footnotesize
\begin{center}
\begin{tabular}{|c|c|c|c|c|c|} \hline
\textbf{Pair Freq.} & \textbf{\# Pairs} & \textbf{Expected \# Same} & \textbf{Expected \% Same} & \textbf{Obs. \# Same} & \textbf{Obs. \% Same} \\ \hline
2 & 33,866 & 16,933.0 & 50.0 & 21,267 & 62.8 \\ \hline
3 & 345 & 86.3 & 25.0 & 128 & 37.1 \\ \hline
4 & 377 & 47.1 & 12.5 & 103 & 27.3 \\ \hline
5 & 28 & 1.8 & 6.3 & 5 & 17.9 \\ \hline
6 & 9 & 0.3 & 3.1 & 1 & 11.1 \\ \hline
\end{tabular}
\end{center}

\caption{Analysis of image pairs that occur multiple times in the training data with the same label.}
\label{tab:freqs}
\end{table}

\begin{figure}[h]
    \centering
    \fbox{\includegraphics[width=0.6\textwidth]{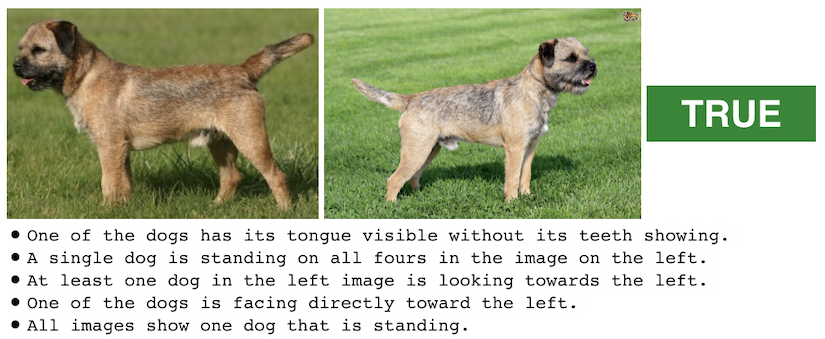}} \\[5pt]
    \fbox{\includegraphics[width=0.6\textwidth]{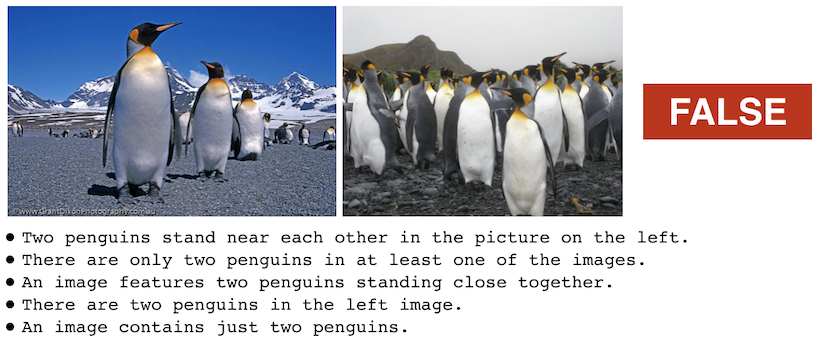}} \\[5pt]
    \fbox{\includegraphics[width=0.6\textwidth]{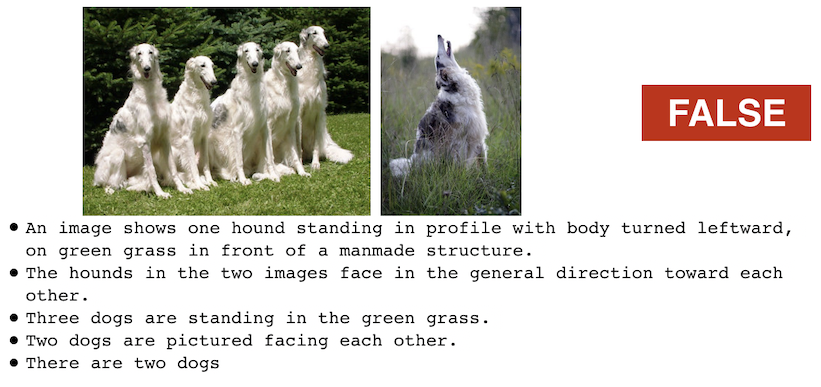}} \\[5pt]
    \fbox{\includegraphics[width=0.6\textwidth]{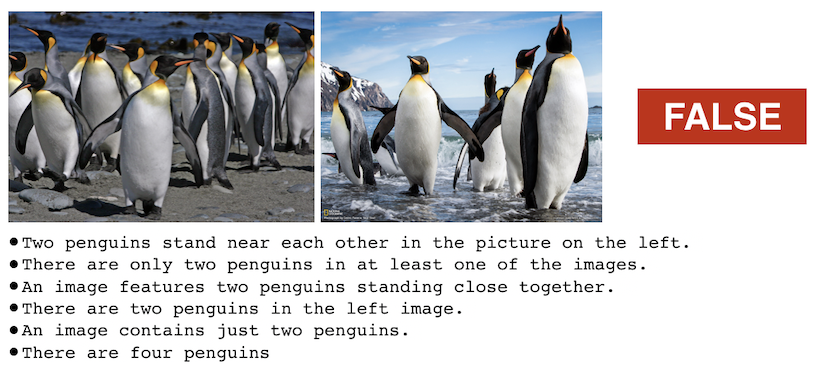}} \\[5pt]
    \fbox{\includegraphics[width=0.6\textwidth]{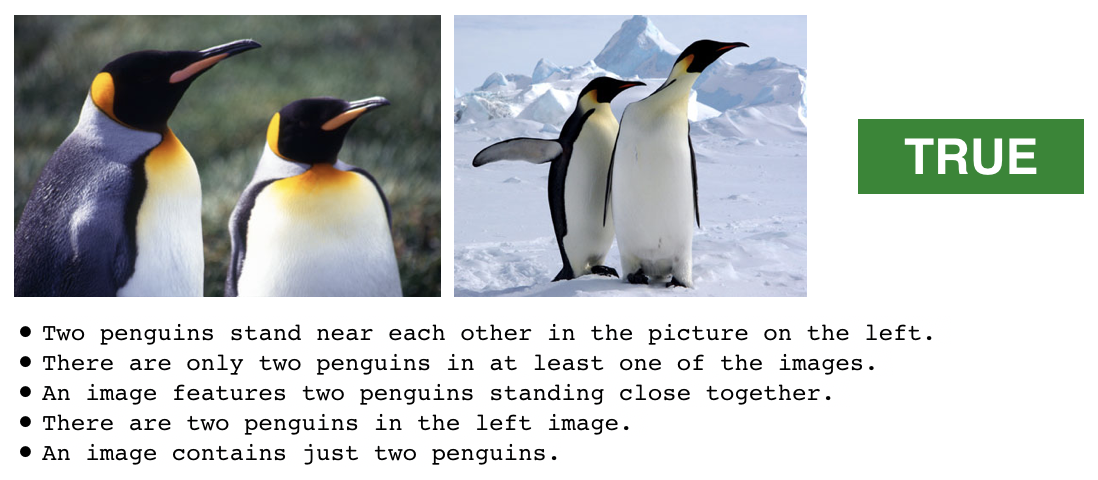}}
\caption{Examples of image pairs in the training data that appear many times with the same label.}
\label{fig:biased_examples}
\end{figure}

We observe  identical labeling  at a proportion (Obs. \% 
Same) much more than expected if assuming labeling events are
independent (Expected \% Same). Figure~\ref{fig:biased_examples} shows several examples of pairs
that appear many times with the same label, along with the label they
were given and the sentences they are paired with.

If the bias is real, we
can try to compute the performance of a model that perfectly learns the bias but
ignores the language. This is a worst-case scenario, and
assumes we can perfectly learn the bias on the evaluation set. We assume
that for all image pairs in the evaluation set, the model outputs the
majority label for this pair for all examples including this pair. For
image pairs that appear once, the model outputs the gold label. If a
pair appears twice with the label \texttt{true}, the model outputs
\texttt{true} for both examples. If a pair appears in three examples
with \texttt{true}, \texttt{false}, and \texttt{true}, the model outputs
\texttt{true}, and gets two out of the three right. We break ties with
\texttt{true}, which is the majority class in the data. We compute this
on the development set and achieve 83.53\% accuracy.

\section{New Evaluation Protocol}\label{sec:newval}

To analyze the extent to which proposed models are taking advantage of any visual bias, 
we can isolate the part of the evaluation set that is
clearly not susceptible to visual bias and test on it. If models perform
on it as well as they do on the general evaluation set, they are not
taking advantage of the bias. We identify all image pairs that appear in
the evaluation data multiple times but with different labels
(\emph{balanced labels}). Because the evaluation sets are much smaller,
the highest frequency of pairs is two. We create new evaluation sets
where each image pair appears twice, once with the label \texttt{true}
and once with \texttt{false}. The new sets are smaller, but still large
enough to evaluate. We only report accuracy on this subset of the data
because consistency~\citep{Goldman:17} changes due to the example selection process and we
want to avoid confusion with the consistency measures of the existing
splits. Similarly, we evaluate on the subset of the data consisting of
pairs that appear more than once with the same label (\emph{unbalanced
labels}).
From the development set, which contains 6,982 examples in total, we obtain
a balanced subset of 2,300 examples and an unbalanced subset of 3,562
examples.

We analyze existing state-of-the-art models on NLVR2.
We evaluate VisualBERT~\citep{Li:19visualbert} and LXMERT~\citep{Tan:19lxmert} 
using the subset of the evaluation data
and show that they largely maintain their reported performance, although
there is a small drop in performance on the balanced set and a small
increase in performance on the unbalanced set.
Table~\ref{tab:sota_results} shows the results of both systems on the balanced and unbalanced subsets of all three evaluation sets.

\begin{table}[t]
\center
\footnotesize
\begin{tabular}{|l|l|c|cc|cc|} \hline
  \textbf{Split}   & \textbf{System} & \textbf{Original Acc.} & \textbf{Balanced Acc.} & \textbf{$\Delta$} & \textbf{Unbalanced Acc.} & \textbf{$\Delta$}  \\ \hline
   \multirow{2}{*}{Development}  & VisualBERT & 67.4 & 66.0 & -1.4 & 69.1 & +1.7 \\ \cline{2-7}
   & LXMERT & 74.9 & 74.0 & -0.9 & 76.4 & +1.5 \\ \hline
   \multirow{2}{*}{Test-P} & VisualBERT & 67.0 & 65.7 & -1.3 & 68.6 & +1.6 \\ \cline{2-7}
   & LXMERT & 74.5 & 73.1 & -1.4 & 76.0 & +1.5 \\ \hline
   \multirow{2}{*}{Test-U} & VisualBERT & 67.3 & 67.3 & +0.0 & 68.2 & +0.9 \\ \cline{2-7}
   & LXMERT  & 76.2 & 76.6 & +0.4 & 76.5 & +0.3 \\ \hline
\end{tabular}
\caption{Performance of VisualBERT~\citep{Li:19visualbert} and LXMERT~\citep{Tan:19lxmert} on the three evaluation sets of NLVR2. For each set, we show the original accuracy on the full set, the performance on the unbalanced and balanced subsets, and the change in performance from the original accuracy when evaluating on the subset of the data. In general, performance decreases slightly when evaluating on the balanced subset and increases slightly when evaluating on the unbalanced subset.}
\label{tab:sota_results}
\end{table}


\section{Conclusion}\label{sec:conclusion}

    We study the potential for visual bias in NLVR2, and show that allowing
the workers to assign labels to image pairs may have introduced some
level of visual bias. We quantify this potential bias, and calculate the
best possible performance for a model relying only on this bias. We also
create new evaluation sets using subset of the original evaluation data.
These new sets are robust to bias, and provide an avenue to test if
models are taking advantage of latent bias. Our evaluation of existing
SOTA models using the new evaluation sets shows that largely they do not
take advantage of latent visual bias and confirm their performance gains.

The key takeaway is the addition of the new evaluation sets for the NLVR2
evaluation. We add these evaluation sets to the NLVR2 data release. We
also recommend evaluating existing models on the original NLVR corpus,
which includes synthetic images, and was constructed to be robust to
both visual and linguistic biases. However, this may be challenging for
models relying on pre-training using natural images.

\section{Acknowledgements}

    We thank Hao Tan and Mark Yatskar for feedback on this note.


\bibliography{main}
\bibliographystyle{apalike}

\end{document}